# MULTI-AUGMENTATION FOR EFFICIENT VISUAL REPRESENTATION LEARNING FOR SELF-SUPERVISED PRE-TRAINING

Van Nhiem Tran[1,2], Chi-En Huang[1], Shen-Hsuan Liu[1,2], Kai-Lin Yang[1], Timothy Ko[2], Yung-Hui Li[1]

[1]AI Research Center, Hon Hai Research Institute, Taipei 114699, Taiwan
[2]Department of Computer Science and Information Engineering, National Central University

[tvnhiemhmus, tigerlittle1, wolfy] @g.ncu.edu.tw
[joseph.ce.huang, kenny.kl.yang, yunghui.li] @foxconn.com

**ABSTRACT**

*In recent years, self-supervised learning has been studied to deal with the limitation of available labeled-dataset. Among the major components of self-supervised learning, the data augmentation pipeline is one key factor in enhancing the resulting performance. However, most researchers manually designed the augmentation pipeline, and the limited collections of transformation may cause the lack of robustness of the learned feature representation. In this work, we proposed **M**ulti-**A**ugmentations for **S**elf-**S**upervised **R**epresentation **L**earning (MA-SSRL), which fully searched for various augmentation policies to build the entire pipeline to improve the robustness of the learned feature representation. MA-SSRL successfully learns the invariant feature representation and presents an efficient, effective, and adaptable data augmentation pipeline for self-supervised pre-training on different distribution and domain datasets. MA-SSRL outperforms the previous state-of-the-art methods on transfer and semi-supervised benchmarks while requiring fewer training epochs. Code available on GitHub[1].*

## 1. INTRODUCTION

Due to the limited available labeled dataset, self-supervised learning has been studied in recent years [1]. Among the major components of self-supervised learning, the data augmentation pipeline is one of the critical factors to affect the resulting performance [2]. Data augmentation pipeline involves geometric transformation such as cropping strategies [3, 4], affine transformation (shear, rotate, translation) [5, 6], appearance colors transformation (color dropping, brightness, contrast, saturation, hue adjustment, etc.) [7, 8]. The pioneering work [2] manually selects each data augmentation policy (type of augmentation and its corresponding magnitude) in the limited searching space (from 8 to 11 transformations only) to build the entire pipeline by investigating the resulting performance on a few combinations of transformations in the searching space. Most previous works [9-11] only fine-tune some transformations based on the pioneering work. Although some of them reveal outstanding results, the limited searching space may constrain the invariant representation that can be learned, which may cause the lack of robustness of the feature representation. Besides, this kind of semi-searching may not guarantee all of the magnitudes have been fully explored. In this work, we proposed MA-SSRL, which applies the various policies searched in the supervised framework to enrich the feature of each image and reveal the truth that the policies searched in the supervised framework may improve the robustness of feature representation in self-supervised learning. The contribution of the proposed method is described as follows:

1. We provided the multiple augmentation strategies, whose policies are fully explored in the large-scale searching space of data augmentation, to improve the robustness of feature representation in self-supervised learning.

2. We applied the random cropping strategies to aggregate the contextual information to improve the robustness of representation in self-supervised learning.

3. We obtain competitive results when transferring representation to multiple natural image datasets on the image classification task.

## 2. RELATED WORK

The data augmentation pipeline is one of the key factors affecting the performance of self-supervised learning. Data augmentation (DAS) for supervised learning has already been studied for several years to fully explore the potential benefits of data augmentation. However, the DAS for self-supervised learning is still a novel topic, so we will mainly review the DAS for supervised learning and mention the latest research progression of DAS for self-supervised learning in this section.

### 2.1 DAS for supervised learning

As the pioneer of the DAS, Ekin et al. proposed AutoAugment [12], which is based on reinforcement learning to find the data augmentation policies by the proximal policy optimization algorithm. To speed up the inference time, Lim

---

[GitHub][1] https://github.com/TranNhiem/MA_SSRL_Tensorflow
[GitHub][1] https://github.com/TranNhiem/MA_SSRL_Pytorch

et al. [13] further proposed Fast AutoAugment to search the augmentation policies by using the Bayesian optimization with Expected Improvement (EI) [14]. Moreover, Cubuk et al.[15] proposed RandAugment, the most efficient way to search the augmentation policies by randomly sampling the transformation and the corresponding magnitude in the large-scale searching space.

## 2.2 DAS for self-supervised learning

The basic idea of DAS for self-supervised learning is derived from SimCLR [2], which systematically investigates the resulting performance of each paired transformation in the searching space based on the linear evaluation. However, the magnitude of each transformation has been frozen during the investigation phase, and the range of magnitude has been constrained. Recently, some pre-text tasks are regarded as the new metric to estimate the resulting performance further to speed up the inference time [16]. For example, better data augmentation should result in better accuracy of the rotation prediction. So, searching the augmentation policies in a large-scale searching space becomes available.

The present work attempts to search the policies for self-supervised learning by investigating the new metric, which is highly correlated with the linear evaluation. However, this metric may cause the bias in estimating the resulting performance in self-supervised learning and become a potential hindrance to an End-to-End framework.

## 3. PROPOSED METHOD

Our ultimate aim is to improve the robustness of the learned feature representation through optimized data augmentation strategy. We introduce **M**ulti-**A**ugmentations for **S**elf-**S**upervised **R**epresentation **L**earning (MA-SSRL) framework shown in Figure 1, which is comprised of two essential parts:

1. The effectiveness of the random uniform distribution cropping strategy enables the encoder to aggregate the contextual information, enhancing the scale-invariancy of the feature embedding.
2. Multi-data augmentation strategy further increases the effectiveness of both supervised and self-supervised learning, which produces more general representation of the natural image and enhances the various invariant properties of the feature embedding.

The detail of each essential part will be described in this section, respectively.

### 3.1. Random cropping with uniform distribution

In general, the entire object may not necessarily appear in the center of the picture. The encoder should extract the feature by focusing on a certain portion of the picture where the object is located. So, the cropping strategy is used to bring the local region matching with the global context in self-supervised learning. In this work, we propose using the random uniform distribution cropping strategy to generate two cropping views, which zoom in to the local region according to the crop ratio. The crop ratio is a dynamically changed parameter by sampling from the random uniform distribution in the range [0.5, 1]. The flowchart of the proposed MA-SSRL is shown in Figure 1, which also shows the scheme for random cropping with uniform distribution (the lower part).

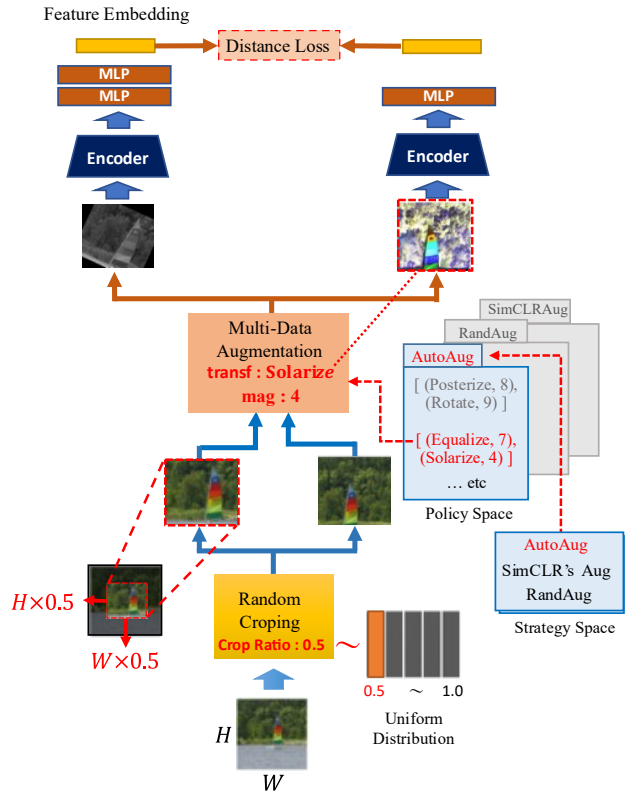

Figure 1. The flowchart of MA-SSRL. The source image is transformed into two views by applying the random cropping strategy, which sets the cropping ratio based on the uniform distribution (bottom). Both cropped views are further augmented by applying the multi-augmentation techniques (center). The encoder (convolution neural network) extracts the invariant feature from both augmented views (middle). The non-linear multi-layer perceptron (MLP) further processes those features and produces corresponding embeddings. Finally, the objective similarity will be measured by contrasting both embeddings of the corresponding view (top).

This work found that random cropping outperforms Inception-style cropping [8], which is applied in most previous works. Detailed information on the Inception-style cropping can be found in the previous work [2]. Since the cropped region will not be constrained by the pre-defined range (from 3/4 to 4/3), it can potentially form a more local view across various scales. We evaluate both cropping strategies on the wide range of augmentation strategies on self-supervised pre-training. The performance comparison between the random cropping versus Inception-style cropping in terms of top-1 accuracy of linear evaluation on top frozen pre-trained representation is presented in Figure 2.

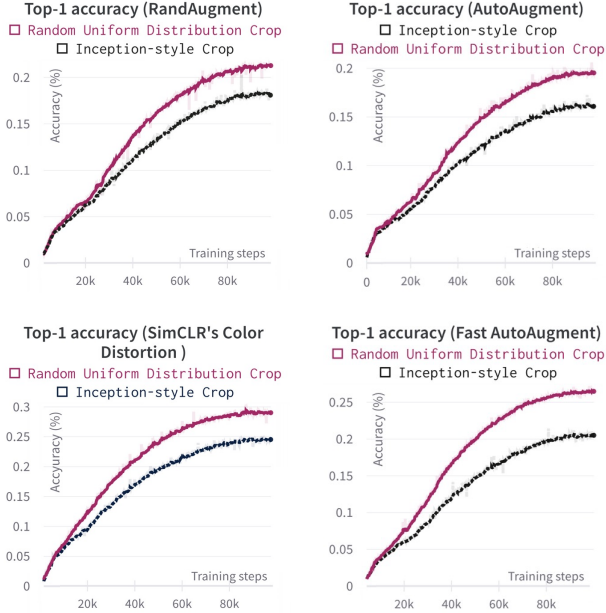

Figure 2. The result on top-1 linear classifier on the subset 100 classes test set (the public validation set of the original ILSVRC2012 ImageNet [17]) with standard ResNet-18 over 100 epochs on 100 classes subset training set of ImageNet (IN).

### 3.2. Multi-data augmentations strategy

Most of the DAS methods try to replicate appearance multiplicity of the natural image by applying the limited kind of transformation in their policy space. In this work, we attempt to extend this idea further and combine the multiple policy spaces of the various augmentation strategies to form an integrated policy space. To explore more general representations of the natural image, we will augment each view by applying the several transformations sampled from the integrated policy space, as illustrated in Figure 1 (the middle part). The various augmented images will force the encoder to embed the information more robustly and focus on the significant feature, which helps the neural network identify the instance given the other views of the same image source. Specifically, the multiple strategy space comprises the three DAS methods: AutoAugment, Fast AutoAugment, and RandAugment. Each of them represents the different paradigms. We verify the impact of supervised augmentation policies using self-supervised pre-training. For doing that, we use one of the effective and efficient supervised augmentation strategies, RandAugment [15]. In our implementation, we further extend the RandAugment searching space and deploy several optimal policies from supervised searched results (number transformation N is sequentially applied, and M corresponds to the magnitude of each transformation). The ablation experiment results are presented in Figure 3. From supervised learning results, the parameter for the optimal performance policy is (N=2 and M=9), of which the performance is similar to the case when (N=1, M=5; N=1 M=10).

Our extended RandAugment search space is defined as:

['AutoContrast', 'Equalize', 'Invert', 'Posterize', 'Solarize', 'Sharpness', 'Color', 'SolarizeAdd', 'ShearX', 'ShearY', 'TranslateX', 'TranslateY', 'rand_brightness', 'rand_contrast', 'rand_saturation', 'rand_hue', 'rand_blur', 'color_drop']

### 3.3. Objective loss function

After augmentation transformation, two augmented views will be generated for each image. The first augmented views (v1) pass through the encoder $f_\theta(h_{v1} = f_\theta(v1)$ where $h \in \mathbb{R}^{H \times W \times D.}$), and project representation by projection network $g_\theta$, where θ is the learned parameters. Another two augmented views (v2) are processed with $f_\xi(h_{v2} = f_\xi(v2)$ where $h \in \mathbb{R}^{H \times W \times D.}$), and $g_\xi$, where $\xi$ is an exponential moving average of θ. The first global is further processed with a predictor network $q_\theta$, with the same architecture as the project network. In the definition of the encoder network and projection network, we re-used the framework structure from BYOL [9], respectively.

$$z_{v1} \triangleq g_\theta \circ q_\theta(h_{v1}) \ ; \ z_{v2} \triangleq g_\xi(h_{v2}) \quad (1)$$

Given the latent representation of the two augmented views ($z_{v1}$ and $z_{v2}$), $\ell_2$-normalization is applied to them, then the cosine distance (the negative cosine similarity) is computed, which is defined as Eq. (2):

$$\mathcal{L}(z_{v1}, z_{v2}) = -\left(\frac{z_{v1}}{\|z_{v1}\|_2} \cdot \frac{z_{v2}}{\|z_{v2}\|_2}\right), \quad (2)$$

where $\|.\|_2$ is $\ell_2$-norm.
We designed the symmetrized loss $\mathcal{L}$ (shown as Eq. (3)) by separately feeding augmented view v1 to the online network and augmented view v2 to the target network and vice versa to compute the loss at each training step. We perform a stochastic optimization step to minimize the symmetrized loss $\mathcal{L}_{symmetrized} = \mathcal{L} + \mathcal{L}^\sim$.

$$\mathcal{L}_{symmetrized} = \tfrac{1}{2}\mathcal{L}(z_{v1}, z_{v2}) + \tfrac{1}{2}\mathcal{L}(z_{v2}, z_{v1}) \quad (3)$$

### 3.4 Evaluation protocol

Following the semi-supervised learning setting in [2, 9, 18], we assess MA-SSRL's performance on the ImageNet ILSVRC-2012 dataset [17]. Initially, we evaluate it in linear, fine-tune evaluation, and semi-supervised settings on ImageNet. Then, we test its transfer performance across a wide range of datasets and tasks. To provide a comparison, we also present scores for a representation trained using the labels from a special ImageNet subset, known as Supervised-ImageNet.

**Linear evaluation on ImageNet:** Following the procedures described in [2, 19], we first evaluate our proposed method by training a linear classifier on top of the frozen representation of the encoder.

**Semi-supervised learning on ImageNet:** In the next stage, we assess the performance obtained by fine-tuning the proposed method with a small subset of the ImageNet training set, using label information. Following [2, 3, 20], we split ImageNet labeled training data into 1% and 10% splits, respectively.

**Transfer learning on natural image datasets:** We assess the effectiveness of our method on other natural image classification datasets to determine if the features learned on ImageNet can be generalized and, thus, useful across different domains or if they are unique to ImageNet. Following the same evaluation protocol as [2, 9, 21], we perform linear evaluation and fine-tuning on the same set of classification tasks they used.

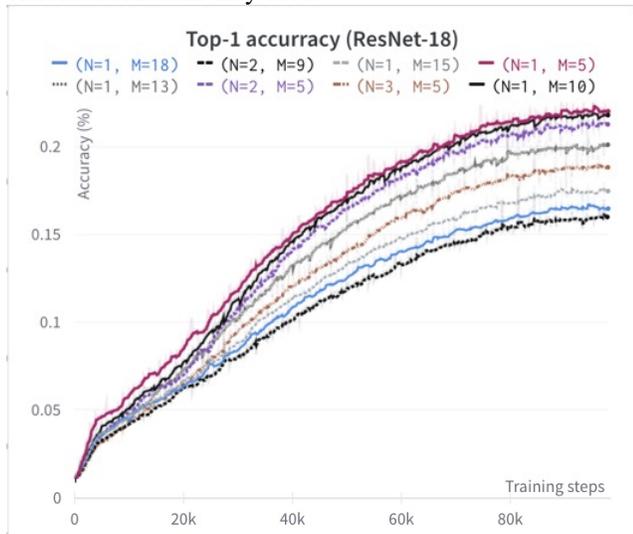

Figure 3. The top-1 linear classifier accuracy of the multiple extended RandAugment policies for self-supervised pretraining on a subset of 100 classes of the training set ImageNet (IN) uses standard ResNet-18.

## 4. EXPERIMENT

### 4.1 Implementation Detail.

**Neural Architecture:** In this study, we implement a standard convolutional residual network architecture. Also, we used different scaling factors of depth (50, 101 layers) and width (from 1x to 2x) in our implementation of ResNets, as in [2, 22]. According to SimCLR [2], the representation of y is projected into a smaller space using a multi-layer perceptron (MLP). In this MLP, the linear layer is followed by batch normalization [23] and rectified linear units (ReLU) [24].

**Optimization:** We apply the LARS optimizer [26] with a cosine decay learning rate schedule [27], without restarts, over 300 epochs. Based on the batch size, we set the base learning rate to 0.3, scaled linearly [28] with the batch size (LearningRate = 0.3 × BatchSize/256). Furthermore, we apply a global weight decay parameter of $1.5\times10^{-6}$ while excluding the biases and normalization parameters from the LARS adaptation and weight decay. We use a batch size of 2048 split over 8 Nvidia A100GPUs. This setup takes approximately 107 hours to train a ResNet-50 (×1).

### 4.2 Linear evaluation and semi-supervised.

We first evaluate MA-SSRL pretraining representation by training a linear classifier on top of the frozen representation of the encoder, and we report top-1 and top-5 accuracies in % on the test set in Table 1. The proposed method obtains 73.8% top-1 accuracy and 91.5% top-5 accuracy using the standard ResNet-50 (×1) with 300 epochs pre-training. It represents a 0.7% and 4.2% advancement over the previous self-supervised state-of-the-art BYOL [9] and SimCLR [2, 9] while reducing the gap with the strong supervised baseline [25].

Then we evaluate the performance obtained after fine-tuning the encoder representation on a small subset of ImageNet's train set using label information. Based on [2, 3, 20], we adhered to the semi-supervised technique and used the same fixed splits of 1% and 10% of ImageNet-labeled training data as in [17]. Table 1 reports both Top-1 and Top-5 accuracies on the test set. MA-SSRL outperforms previous self-supervised frameworks [2, 9, 26] and achieves a significant improvement of 56.3% and 69.1% top-1 accuracy, which is a +3% and +7.9% improvement over the previous strong baseline BYOL [9] and SimCLR [2] of 1% ImageNet labeled IN.

### 4.3 Transfer Learning on different natural image datasets

We study MA-SSRL's learned representation generalization and robustness by transferring it to other natural image dataset classifications. As a result, our representation transfers well across various datasets. We evaluated the MA-SSRL's representation learning of neural network encoder on linear classification and fine-tuning following the evaluation protocol [2, 9, 18, 21].

Table 1. The results of linear evaluation and semi-supervised learning with a fraction (1% and 10% ImageNet labeled IN) in top-1 and top-5 accuracy (%) use the ResNet-50 encoder.

| Method | Linear Evaluation | | Semi-Supervised | | | |
|---|---|---|---|---|---|---|
| | Top-1 | Top-5 | Top-1 | | Top5 | |
| | | | 1% | 10% | 1% | 10% |
| Supervised IN [2] | 76.5 | - | 25.4 | 56.4 | 48.4 | 80.4 |
| PIRL [26] 800-epochs | 63.6 | - | - | - | 57.2 | 83.8 |
| SimCLR [2] 1000-epochs | 69.3 | 89.0 | 48.3 | 65.6 | 75.5 | 87.8 |
| MoCo v2 [27] 1000-epochs | 71.1 | - | - | - | - | - |
| BYOL [9] 1000-epochs | 74.3 | 91.6 | 53.2 | 68.8 | 78.4 | 89.0 |
| BYOL [9] 300-epochs | 72.4 | 90.7 | - | - | - | - |
| BYOL (repo) 300-epochs | 72.8 | 91.0 | 52.4 | 67.7 | 77.9 | 88.5 |
| **MA-SSRL** (ours) 300-epochs | 73.8 | 91.5 | 56.3 | 69.1 | 80.4 | 89.1 |

Our MA-SSRL framework outperforms several strong baseline self-supervised frameworks, including SimCLR [2], BYOL [9], on 6 over 6 different natural distributions image datasets (namely Food-101 [28], and CIFAR-100 [29], the SUN397 scene dataset [30], Stanford Cars [31], the Describable Textures Dataset (DTD) [32], Oxford-IIIT Pets[33]). On datasets without a test set or validation, we use the validation examples as the test set or hold out a subset of the training examples in [9]. MA-SSRL's representation can perform well in largely different distributions, as shown in Table 2.

## 5. CONCLUSION

We introduce the MA-SSRL framework, the efficient augmentation pipeline for self-supervised representation learning. MA-SSRL's augmentation pipeline is dynamic and adaptable for the self-supervised pretraining stage in different image distribution and domain adaption datasets by leveraging the multiple supervised augmentations searched policies. As a result, MA-SSRL's representation learning produces competitive results compared to previous state-of-the-art methods [2, 9, 10, 34] on semi-supervised and transfer learning on various benchmarks.

Table 2. Transfer learning of the MA-SSRL representation trained on ImageNet (IN) uses the standard ResNet-50 backbone (IN).

| Methods | Food101 | CIFAR 100 | Cars | DTD | Sun397 | Pets |
|---|---|---|---|---|---|---|
| *Linear evaluation:* | | | | | | |
| **MA-SSRL** (ours) 300-epochs | **76.0** | **78.9** | **57.7** | **73.8** | **63.8** | **84.3** |
| BYOL (repo) 300-epochs | 72.2 | 75.4 | 46.3 | 72.9 | 62.5 | 82.4 |
| BYOL [9] (1000 epochs) | 75.3 | 78.4 | 67.8 | 75.5 | 62.2 | 90.4 |
| SimCLR[2] 1000 -epochs | 68.4 | 71.6 | 50.3 | 74.5 | 58.8 | 83.6 |
| *Fine-tune:* | | | | | | |
| **MA-SSRL** (ours-300 epochs) | **85.4** | **85.8** | 84.3 | 70.4 | **63.5** | 80.5 |
| BYOL (repo 300-epochs) | 85.1 | 83.3 | 86.1 | 71.3 | 62.3 | 85.3 |
| BYOL [9] (1000 epochs) | 88.5 | 86.1 | 91.6 | 76.2 | 63.7 | 91.7 |
| SimCLR[2] (1000 epochs) | 88.2 | 85.9 | 91.3 | 73.2 | 63.5 | 89.2 |

## 6. REFERENCES

1. Jing, L. and Y. Tian, *Self-Supervised Visual Feature Learning With Deep Neural Networks: A Survey.* IEEE Transactions on Pattern Analysis and Machine Intelligence, 2021. **43**: p. 4037-4058.
2. Chen, T., et al., *A Simple Framework for Contrastive Learning of Visual Representations.* ArXiv, 2020. **abs/2002.05709**.
3. Hjelm, R.D., et al., *Learning deep representations by mutual information estimation and maximization.* ArXiv, 2019. **abs/1808.06670**.
4. HÈnaff, O.J., et al., *Data-Efficient Image Recognition with Contrastive Predictive Coding.* ArXiv, 2020. **abs/1905.09272**.